\pdfoutput=1

\documentclass[11pt]{article}
\usepackage[toc,page]{appendix}
\usepackage[preprint]{acl}

\usepackage{times}
\usepackage{latexsym}

\usepackage[T1]{fontenc}

\usepackage[utf8]{inputenc}

\usepackage{microtype}

\usepackage{inconsolata}

\usepackage{graphicx}

\title{Can CLIP Count Stars? An Empirical Study on Quantity Bias in CLIP}

\author{Zeliang Zhang\ , Zhuo Liu\ , Mingqian Feng, Chenliang Xu\\\
Department of Computer Science, University of Rochester\\\\
\footnotesize{\texttt{ \{zeliang.zhang, zhuo.liu, susan.liang,chenliang.xu\}@rochester.edu}}\\
}

\begin{document}
\maketitle
\begin{abstract}
CLIP has demonstrated great versatility in adapting to various downstream tasks, such as image editing and generation, visual question answering, and video understanding. However, CLIP-based applications often suffer from misunderstandings regarding user intent, leading to discrepancies between the required number of objects and the actual outputs in image generation tasks. In this work, we empirically investigate the quantity bias in CLIP. By carefully designing different experimental settings and datasets, we comprehensively evaluate CLIP's understanding of quantity from text, image, and cross-modal perspectives. Our experimental results reveal a quantity bias in CLIP embeddings, impacting the reliability of downstream tasks.
\end{abstract}

\section{Introduction}
The Contrastive Language-Image Pre-Training (CLIP) model~\citep{radford2021learning}, trained on large-scale image-text pairs, has shown significant success in various downstream vision-language tasks, including editing~\citep{guerrero2024texsliders,michel2024object}, generation~\citep{ganz2024clipag,liu2024isotropic3d}, and quality evaluation~\citep{hong2024s,deng2024seeing}. It is crucial to maintain a reliable CLIP model at the core to ensure the development of trustworthy applications built upon it.

However, several factors potentially hinder the interpretability and trustworthiness of CLIP, including the black-box nature of the learning process, uneven distributions of the training data, and the difficulty in accurately learning specific data distributions. Such issues may lead to unintended systematic errors like spurious correlations~\citep{sagawa2020investigation} and subgroup biases~\cite{zhang2024discover}. These drawbacks not only degrade CLIP's performance in learning reliable latent representations for image and text translation, but also pose a risk of propagating unexpected biases to models that utilize CLIP for downstream tasks, thereby resulting in more challenging bugs to fix~~\citep{tanjim2024discovering}.


\begin{figure}
    \centering
    \includegraphics[width=\linewidth]{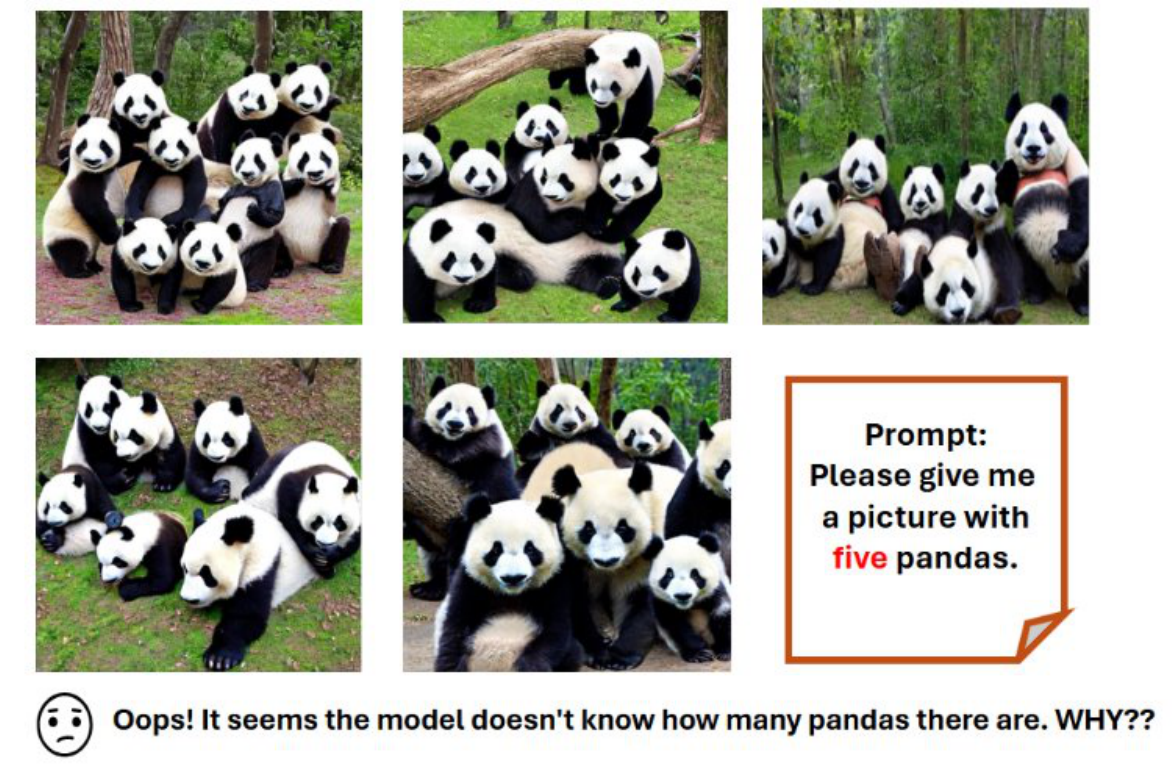}
    \caption{Existing models often show the quantity bias in different tasks. In this example, CLIP-based stable-diffusion model mostly generates a picture with seven pandas, while we only prompt the five.}
    \label{fig:quantity_bias_teaser}
\end{figure}

For instance in Fig. \ref{fig:quantity_bias_teaser}, when using Stable-Diffusion (SD)~\citep{rombach2022high}, which leverages the relationship between image and text learned from CLIP~\citep{ding2024clip}, to generate an image of five pandas, most of the output consistently depicts seven pandas. We query the reason from the foundational CLIP model rather than the SD model at the top layer, as debugging becomes significantly more difficult and challenging if the foundational models are already biased. This intriguing phenomenon raises an important question: \textit{Can CLIP count stars}?

To address this question, our work investigates the quantity bias in CLIP. Specifically, we empirically evaluate CLIP at three levels: text, image, and cross-modal interactions. For each level, we set up tasks of varying difficulty to ensure a comprehensive evaluation.  Our findings highlight the need for addressing these biases to enhance the reliability and effectiveness of CLIP in real-world applications.

We summarize our findings as follows,
\begin{itemize}
    \item CLIP models can not understand the concept of quantity in text-only, image-only, or cross-modality contexts.
    \item CLIP can distinguish the semantic difference between different quantity words but fails to compare them effectively.
    \item CLIP can not effectively find the semantic difference between images with different number of same or similar objects, also leading to the failure of quantity identification. 
\end{itemize}



\section{Related work}
There have been many efforts working on the model bias. \citet{kotek2023gender} investigate the behavior of large language models on gender bias. \citet{liu2022quantifying} measure the political bias in language models and propose a reinforcement learning-based method to mitigate the bias. \citet{zhang2024discover} identify the existence of subgroup bias in image classifiers and use a supervised decomposition method to discover unknown bias from the joint information from the model and inputs. \citet{hosseini2018assessing} find the shape bias learning by convolutional neural networks. \citet{khayatkhoei2022spatial} discover generative models can easily learn the spatial bias from the data. \citet{heinert2024reducing} and \citet{honig2024star} research on the texture bias of deep learning models and downstream tasks. 

Different from previous task-specific and application-driven studies on model bias, we study the bias of the embedding in CLIP. There are mainly two reasons motivating us: \underline{First}, various studies have identified that there is fruitful semantic information in the embedding, where the existence of bias could have a great impact on the whole model. \underline{Second}, as CLIP serves as a vision-language foundation model in many downstream tasks and model developments, it can help us understand the model behavior by studying the bias issue of the used foundation model.  In this work, we study the existence of quantity bias from the CLIP embedding level for a better understanding of the failure of generative models.

\section{Quantity Bias in CLIP}

\subsection{Experiment design}

\noindent \textbf{Overview}. We study the quantity bias in CLIP at two levels: uni-modal and multi-modal. In each modal capacity, we examine the concept association between different quantitative nouns and comparative descriptions. Then, we investigate the quantity bias in cross-modal capacity.  Furthermore,  we use CLIP-based retrieval results to reflect the impact of quantity bias in CLIP.  

\noindent \textbf{Models}. Nine CLIP models are evaluated in our study, including RN-50~\citep{he2016deep}, RN-101, RN-50x4, RN50x16, RN50x64, ViT-B/32~\citep{han2022survey}, ViT-B/16, ViT-L/14,  and ViT-L/14@336px. We get the pre-trained models from the CLIP library~\citep{radford2021learning}.

\noindent \textbf{Datasets}. We manually construct the dataset for the quantity bias study. For the text modality, we create various quantity-related nouns such as 'zero,' 'three,' and 'hundreds.' For the image modality, we generate images with different numbers of circles, where the positions of the circles are randomly sampled. To better reflect CLIP's knowledge of quantity, we use descriptive nouns to benchmark various detailed quantity nouns, such as 'many,' 'fewer,' and 'lots of.' We study the quantity bias at the embedding level, which encodes the rich semantic information learned by the CLIP model.

\noindent \textbf{Evaluation metric}.  Following previous studies, we use the inner product as the similarity score between embedding to evaluate the semantic correlation between different words and concepts. 


\begin{figure}
    \centering
    \includegraphics[width=\linewidth]{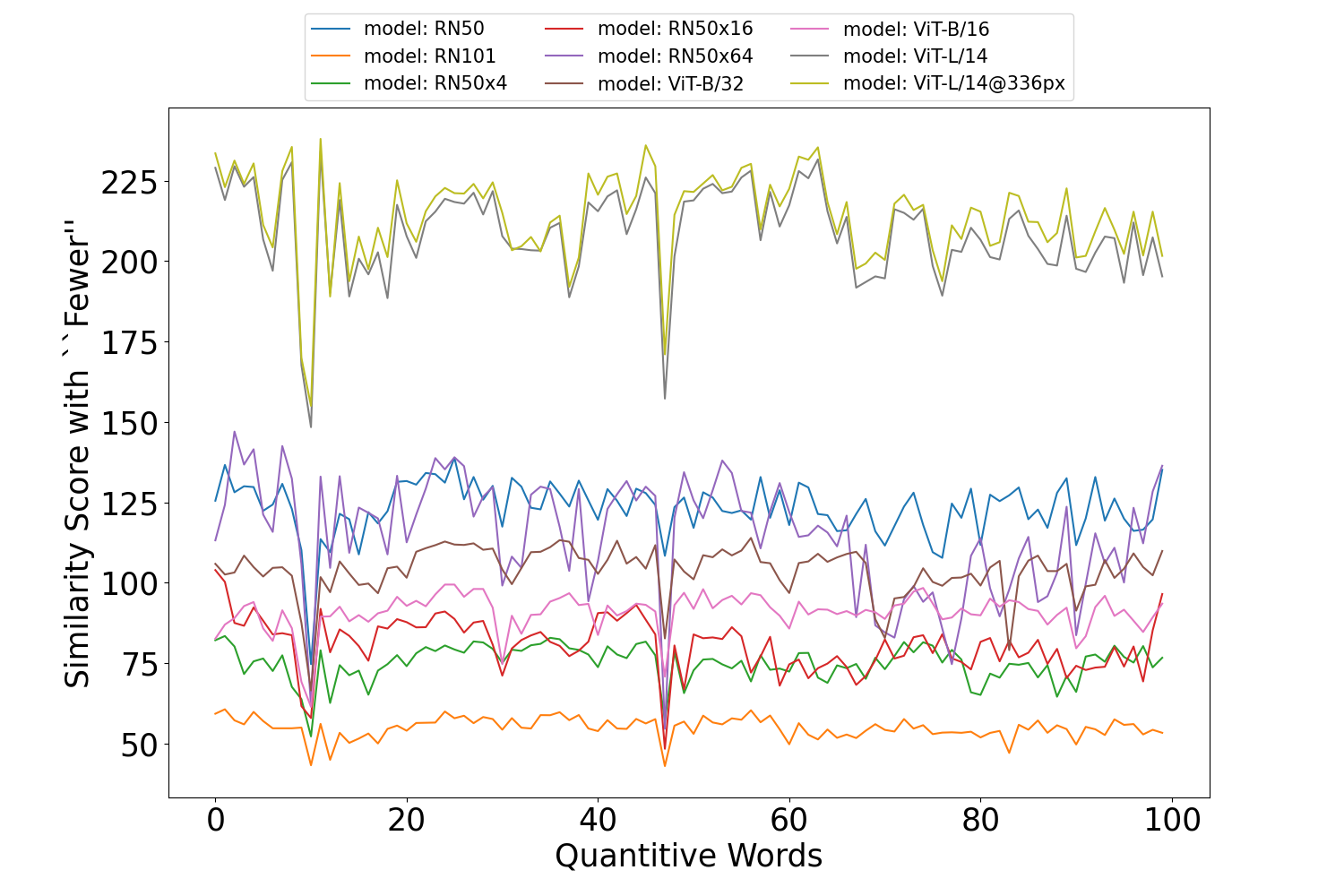}
    \caption{How ``fewer'' for selected different words.}
    \label{fig:text_fewer}
\end{figure}

\begin{figure}
    \centering
    \includegraphics[width=\linewidth]{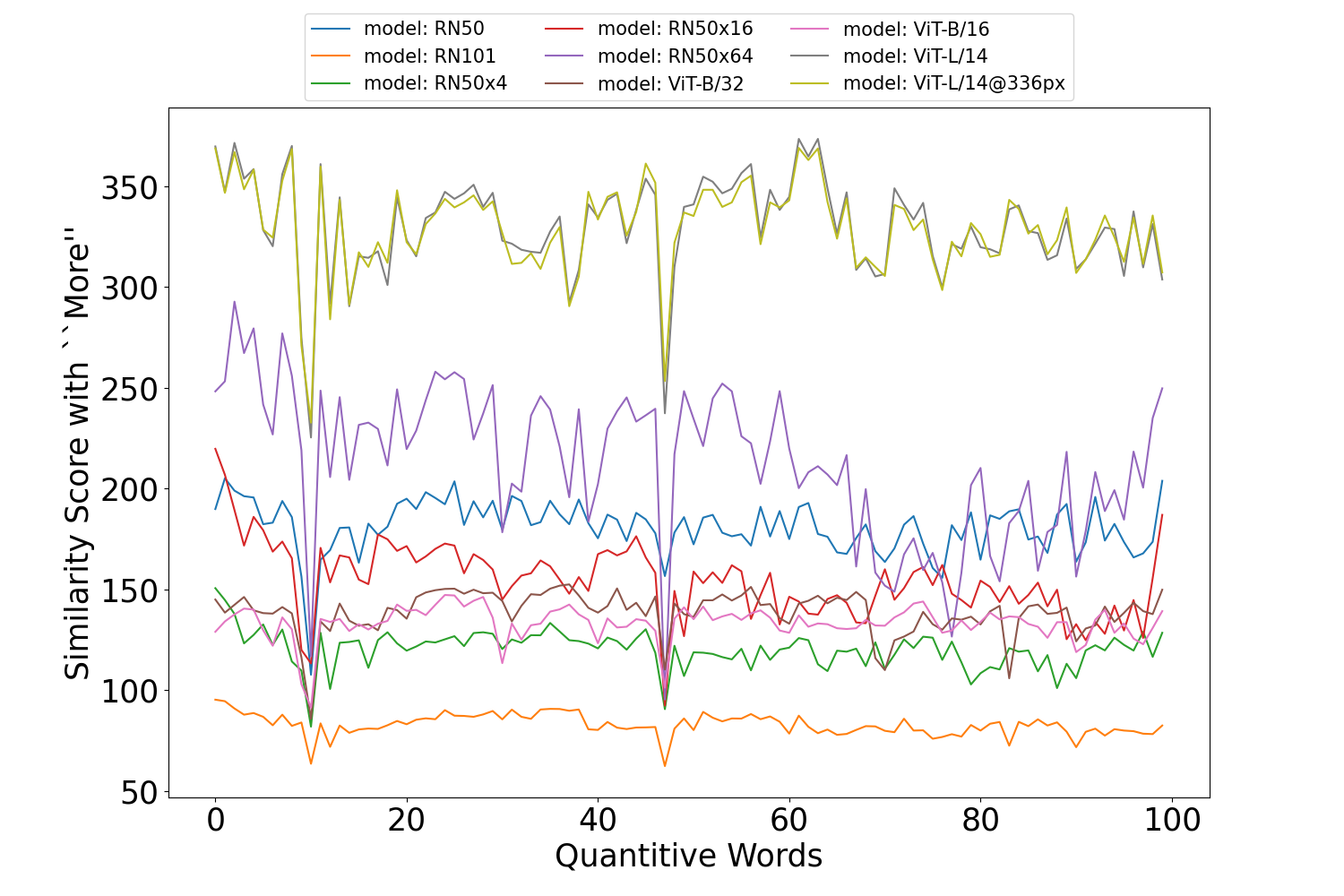}
    \caption{How ``more'' for selected different words.}
    \label{fig:text_more}
\end{figure}

\subsection{Evaluation on the uni-modal capacity}

Texts are mostly used in CLIP-based downstream tasks, which are more intuitive for human understanding. While our human can easily describe what is less and what is more, here, we question whether the CLIP also knows.

We build a small dataset containing 25 specific quantity nouns, ranging from 0 to 100. Two benchmark words for comparable description are used: 'fewer' and 'more.' Through this study, we aim to understand how CLIP interprets the concepts of 'few' and 'more.' We report the results of similarity score for the ``fewer'' and ``more'' study in Fig. \ref{fig:text_fewer} and Fig. \ref{fig:text_more}, respectively.  The statistical results reveal some interesting phenomena. 

First, \textit{CLIP cannot distinguish different quantity words well}. As the quantitative word becomes smaller or larger, the similarity score with 'fewer' or 'more' doesn't decrease or increase gradually. Additionally, we can clearly see that 'zero' demonstrates the highest similarity with both 'fewer' and 'more.' This could be the first evidence showing that CLIP cannot count and understand quantitative words well.

\begin{figure}
    \centering
    \includegraphics[width=\linewidth]{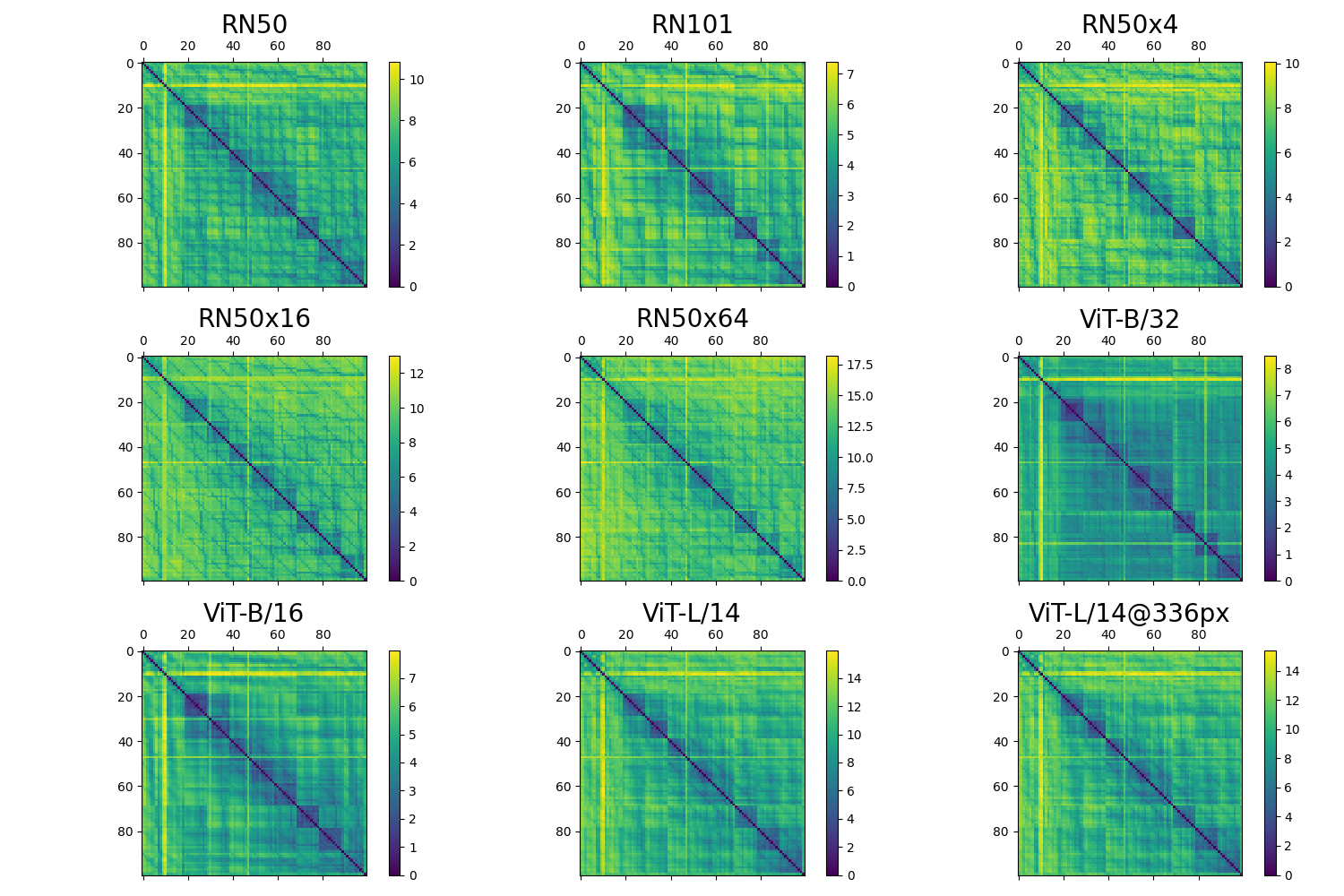}
    \caption{Distance between different quantity words.}
    \label{fig:text_pairdistance}
\end{figure}

Second, \textit{the quantity bias in the text modality is shared across different models}. We surprisingly find that the change in similarity scores for different models follows a similar trend, \textit{i.e.}, the peaks and troughs, while differing in magnitude. For example, the values of most models at around 'fifth' and 'thirteenth' exhibit the maximum and minimum similarity scores, respectively, when compared with both 'fewer' and 'more.' These results indicate that the quantity bias is a systematic error, not a model-specific error.

To have a better understanding on the embedding of quantitive words and quantity bias, we compute the pairwise distance in $\mathcal{L}_{2}$ norm between different words. The results are  shown in Fig. \ref{fig:text_pairdistance}. With a darker color indicating a smaller distance, we can see that closely neighboring words present high similarity at the embedding level, while distant words demonstrate low similarity. On the one hand, \textit{the high semantic similarity of closely neighboring words makes quantity comparison tricky.} On the other hand, the semantic sensitivity of the distance in text embedding causes rapid and discontinuous changes in quantity comparison. Additionally, the words around 'ten' to 'thirteenth' show a large distance in the embedding space compared to other words, which is consistent with previous finding in Fig. \ref{fig:text_fewer} and Fig. \ref{fig:text_more}.

\begin{figure}
    \centering
    \includegraphics[width=\linewidth]{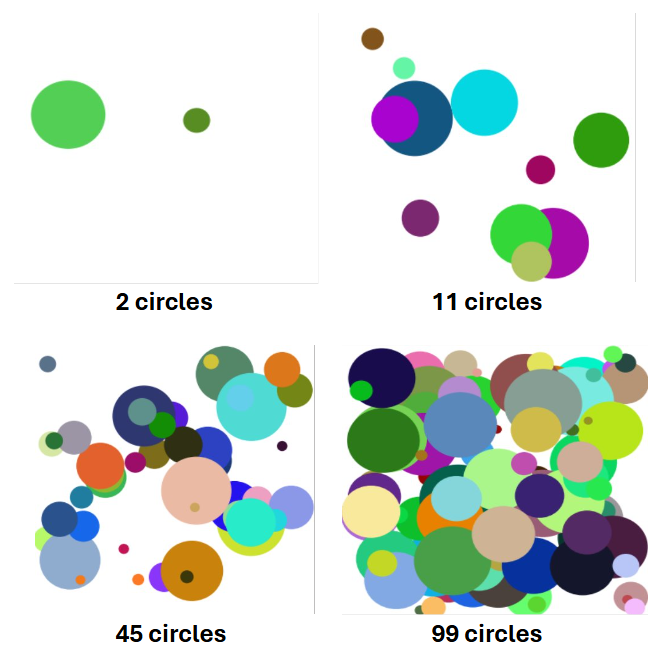}
    \caption{Examples of the images with different number of circles.}
    \label{fig:example_circle}
\end{figure}

Moving to the image domain, with lacking explicit signal for quantity comparison described by pure image modality, we only compute the pair wise distance for images with different number circles. The images are randomly generated with given number of circles. We present some examples in Fig. \ref{fig:example_circle}.

\begin{figure}
    \centering
    \includegraphics[width=\linewidth]{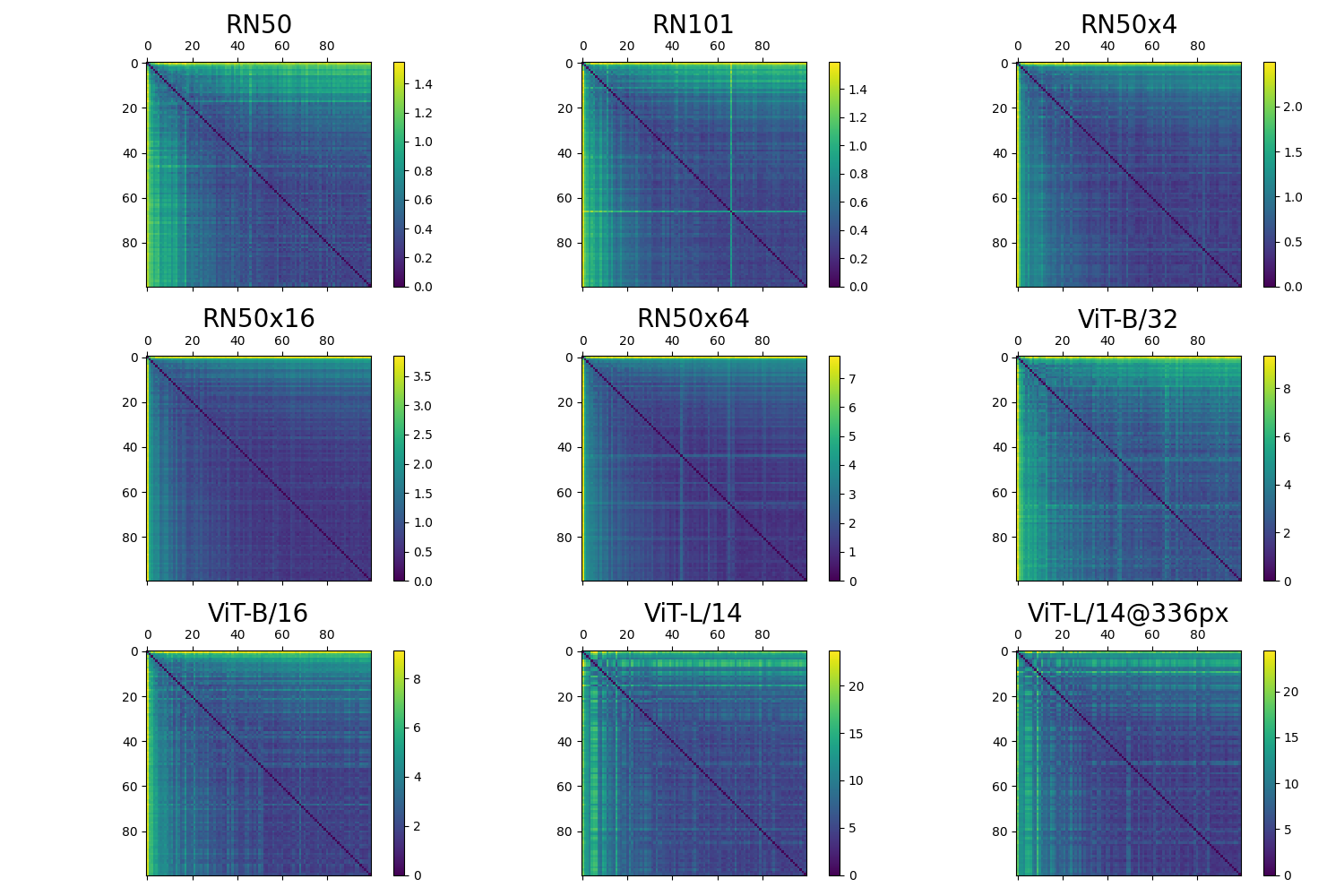}
    \caption{Distance between images with different number of objects.}
    \label{fig:image_pairdistance}
\end{figure}

Unlike in the text domain, as shown in Fig. \ref{fig:image_pairdistance}, different images show low distance in the embedding space to each other, indicating the difficulty in differentiating them at the quantity level.

\paragraph{Take-home message} CLIP cannot understand the concept related to quantity in either the text or image modality, though for different reasons. In the text domain, there is high similarity between closely neighboring quantities, but a large semantic difference with distant quantity words. The irregular and noncontinuous changes between continuous quantity words make comparison difficult, leading to confusion with 'fewer' and 'more.' Conversely, in the image domain, images with different numbers of circles show high semantic similarity, making it difficult to differentiate them based on semantic differences. These factors lead to CLIP's failure in understanding quantity.

\subsection{Evaluation on the multi-modal capacity}

We further evaluate the quantity bias  in multi-modal capacity of CLIP models. We use the quantities comparison words ``fewer'' and ``more'' to evaluate the figures with different number of circles introduced before.

\begin{figure}
    \centering
    \includegraphics[width=\linewidth]{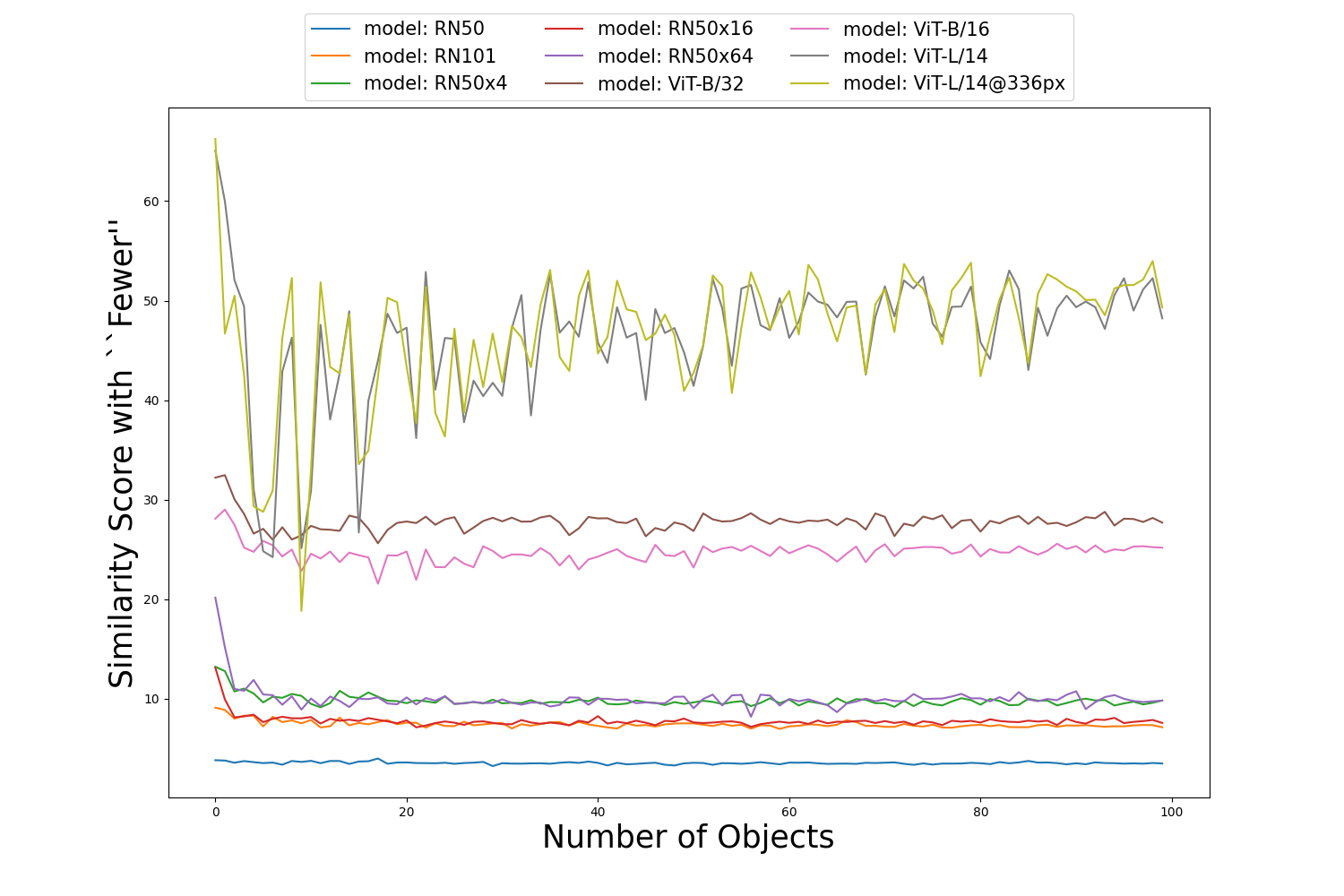}
    \caption{How ``fewer'' for generated objects.}
    \label{fig:image_fewer}
\end{figure}

\begin{figure}
    \centering
    \includegraphics[width=\linewidth]{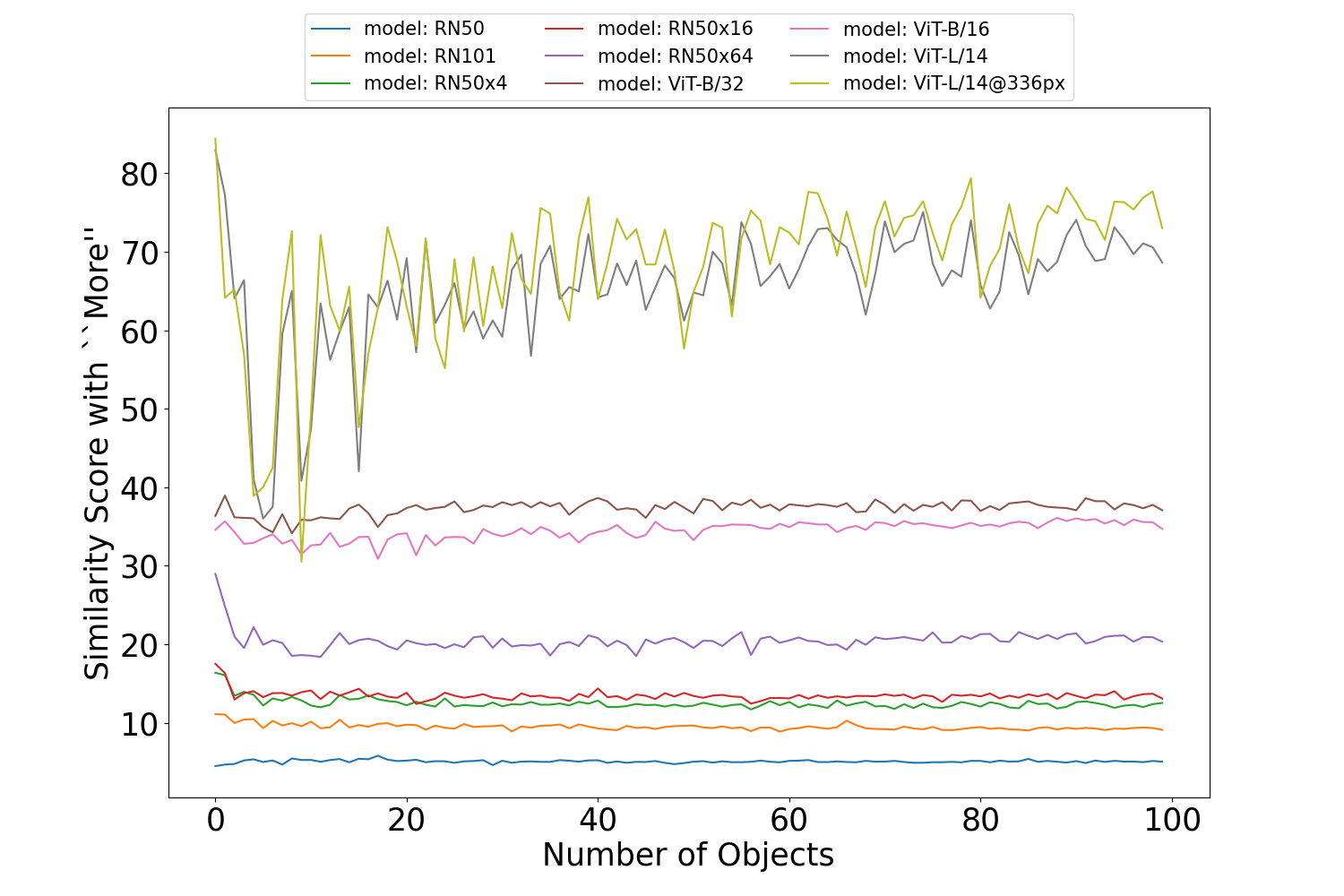}
    \caption{How ``more'' for generated objects.}
    \label{fig:image_more}
\end{figure}

We report the similarity comparison results in Fig. \ref{fig:image_fewer} and Fig. \ref{fig:image_more}. It can be seen that most models cannot distinguish different images at the embedding level for the 'fewer' or 'more' concepts, with the similarity scores remaining smooth at a low level. Although the two ViT-L models show a large difference in word embedding between different images, they also share the same trend for the 'fewer' and 'more' concepts. This indicates that these two CLIP models learn the difference between different numbers of circles due to larger model capacity but still fail to understand quantity.

\subsection{Discussion}
\noindent \textbf{Why this happens}? We argue that two factors contribute to the ineffective learning of the quantity concept in CLIP models. First, quantity-related data are heavily limited. There are not enough data containing explicit quantity information for model learning. Many quantity words and quantity-related visual information are not considered in the model learning process, effectively making them out-of-distribution data. Second, there is a critical technical flaw in contrastive learning. While contrastive learning enables the model to learn a unified representation of different modalities, it overlooks many attributes of the inputs. For example, when contrastive learning requires CLIP to map images with circles and the text description 'circles' to the same latent space, it overlooks the comparison of the shape and number of circles in text and image modalities, leading to the existence of quantity bias in the well-trained models.

\noindent \textbf{How to mitigate such bias}? Based on our findings, we think there are two set of strategies for bias mitigation, including the data-centric and model-centric strategy.

\begin{itemize}
    \item \textit{Data-centric}: It's crucial to construct high-quality multimodal data for the training and fine-tuning of foundational models like CLIP. Instead of simply pairing images with their names for contrastive learning, the textual descriptions should include more detailed attributes such as quantity, color, and shape. This enriched information can help distinguish the embeddings from each other in the latent space, thereby reducing embedding bias and minimizing confusion for downstream task models.
    \item \textit{Model-centric}: Mitigation strategies should be tailored to specific real-world applications. For instance, addressing the counting problem highlighted in our paper, while it may be time- and computation-intensive to modify the foundational model, fine-tuning downstream task models like Stable-diffusion with carefully designed prompts, such as "many" and "fewer," can be more practical. Additionally, developers can include a regularization term that distinguishes and group different sets of quantitative words, like "one," "two" for "smaller", and "hundreds," "thousands" for "larger". This approach encourages the model to learn and differentiate quantitative concepts more effectively.
\end{itemize}

\section{Conclusion}

In this work, we study an interesting problem: \textit{Can CLIP count stars?} Through extensive empirical studies on different modalities, we conclude that the CLIP models cannot understand the concept of quantity well. In the future, we will delve deeper into this quantity bias and design novel methods for efficient bias mitigation.

\section*{Limitations}
CLIP is one of the most popular foundation models used in generation tasks (e.g., the development of Stable Diffusion), which motivates us to study a variety of CLIP models with different vision and text backbones in this short paper. To ensure a controlled examination of the variable in question—specifically, the number of objects in both visual and textual modalities—we employed manually constructed datasets to evaluate the quantity bias of CLIP in this preliminary research. However, real-world data presents more diversity and complexity, which were not fully captured in the simulations of this study. Future research should include more extensive results from real-world datasets and evaluate a broader range of vision-language models to provide a more comprehensive assessment.

\bibliography{custom}

\begin{appendices}

\setcounter{table}{0}   
\setcounter{figure}{0}
\renewcommand{\thetable}{A\arabic{table}}
\renewcommand{\thefigure}{A\arabic{figure}}

\section{More failure examples of CLIP-guided Stable-diffusion for image generation}

\begin{figure}
    \centering
    \includegraphics[width=0.7\linewidth]{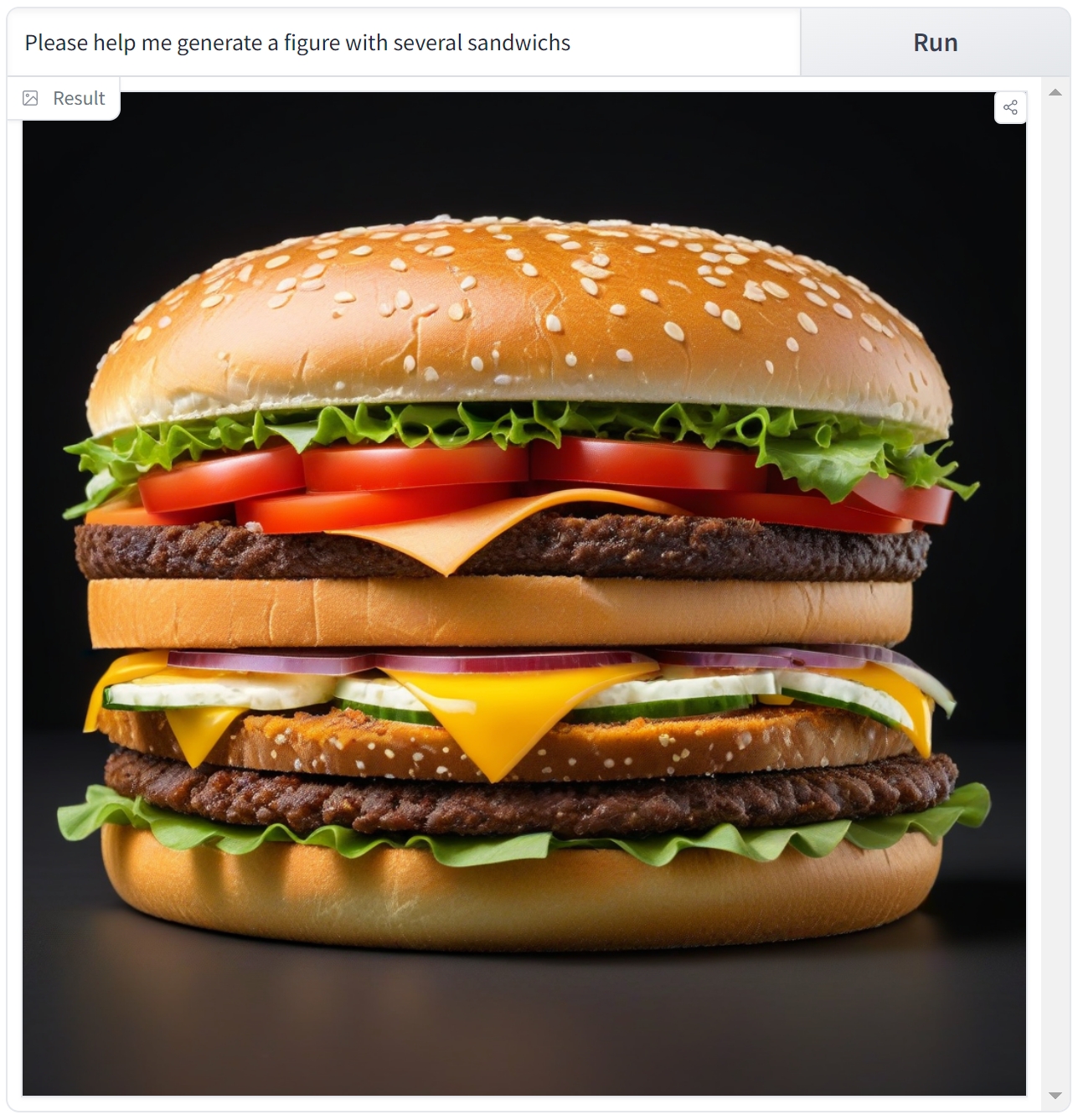}
    \caption{Prompt: Please help me generate a figure with \textbf{several} sandwiches.}
    \label{fig:sandwich}
\end{figure}

\begin{figure}
    \centering
    \includegraphics[width=0.7\linewidth]{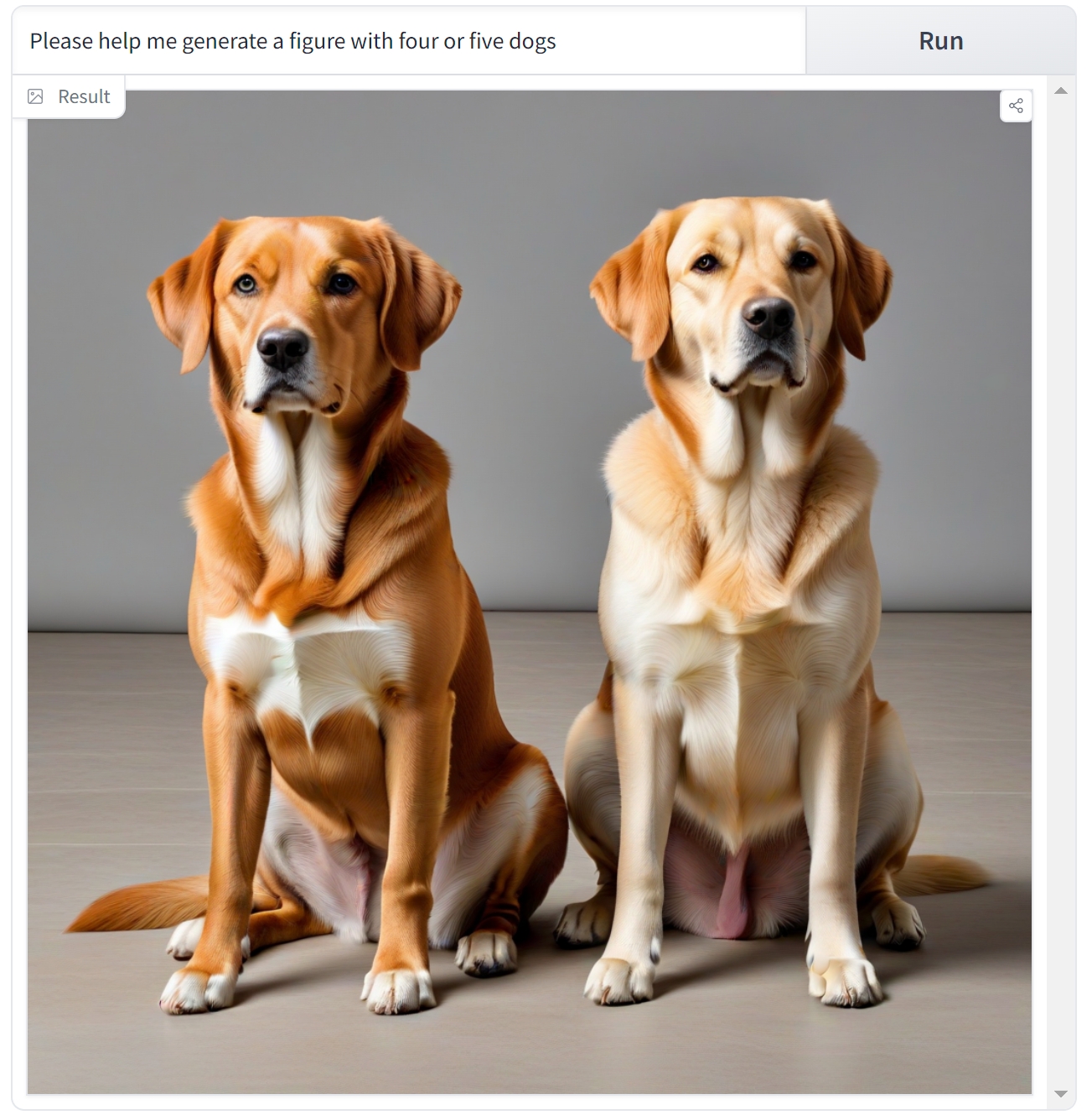}
    \caption{Prompt: Please help me generate a figure with \textbf{four} dogs.}
    \label{fig:sandwich}
\end{figure}

\begin{figure}
    \centering
    \includegraphics[width=0.7\linewidth]{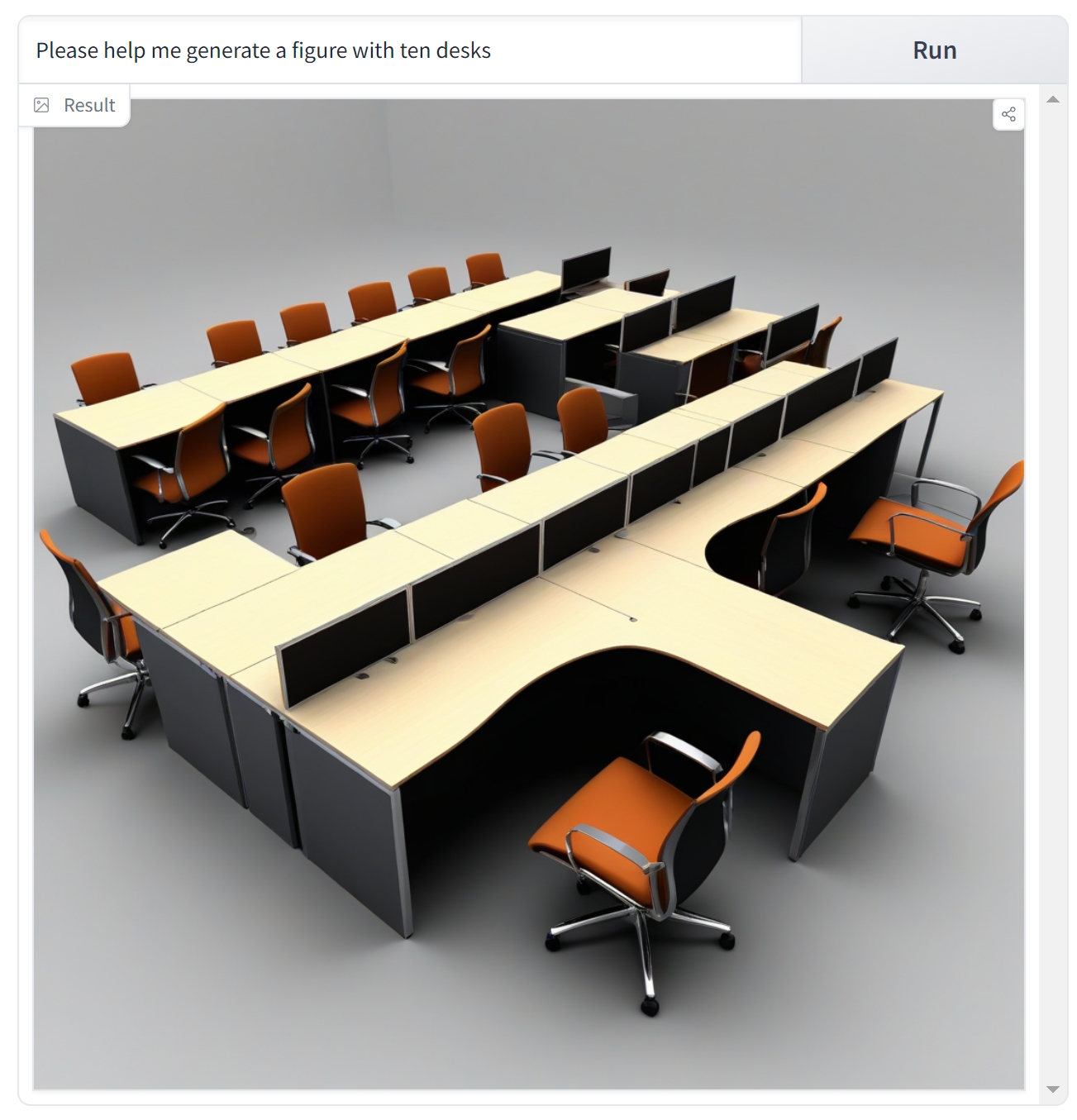}
    \caption{Prompt: Please help me generate a figure with \textbf{ten} desks.}
    \label{fig:sandwich}
\end{figure}

\end{appendices}

\end{document}